\documentclass{article}

\usepackage[margin=1in]{geometry}
\usepackage{graphicx}
\usepackage{amsmath}
\usepackage[table,xcdraw]{xcolor}
\usepackage{amssymb}
\usepackage{hyperref}
\usepackage{subcaption}
\usepackage[toc,page]{appendix}
\usepackage{cite}
\usepackage{multirow}
\usepackage{authblk}
\usepackage{tikz}
\usetikzlibrary{positioning, arrows.meta, chains, fit, shapes.geometric, calc}

\providecommand{\keywords}[1]{\textbf{\textit{Keywords---}} #1}

\begin{document}

\title{Deep Learning using Rectified Linear Units \\ \normalsize A rectified version of the 2018 paper}

\author{Abien~Fred~Agarap\\\small\texttt{abien.agarap@dlsu.edu.ph}}
\affil{College of Computer Studies\\De La Salle University\\Manila, Philippines}
\date{}

\maketitle

\begin{abstract}
The Rectified Linear Unit (ReLU) is a foundational activation function in artficial neural networks. Recent literature frequently misattributes its origin to the 2018 (initial) version of this paper, which exclusively investigated ReLU at the classification layer. This paper formally corrects the citation record by tracing the mathematical lineage of piecewise linear functions from early biological models to their definitive integration into deep learning by \cite{nair2010rectified}. Alongside this historical rectification, we present a comprehensive empirical comparison of the ReLU, Hyperbolic Tangent (Tanh), and Logistic (Sigmoid) activation functions across image classification, text classification, and image reconstruction tasks. To ensure statistical robustness, we evaluated these functions using 10 independent randomized trials and assessed significance using the non-parametric Kruskal-Wallis $H$ test. The empirical data validates the theoretical limitations of saturating functions. Sigmoid failed to converge in deep convolutional vision tasks due to the vanishing gradient problem, thus yielding accuracies equivalent to random probability. Conversely, ReLU and Tanh exhibited stable convergence. ReLU achieved the highest mean accuracy and F1-score on image classification and text classification tasks, while Tanh yielded the highest peak signal to noise ratio in image reconstruction. Ultimately, this study confirms a statistically significant performance variance among activations, thus reaffirming the necessity of non-saturating functions in deep architectures, and restores proper historical attribution to prior literature.
\end{abstract}
\keywords{Activation functions, artificial neural networks, deep learning}

\section{Introduction and Related Works}
A neural network is a machine learning model that approximates a function $f(\vec{x})$ showing the relationship between the dataset features $\vec{x}$ and labels $\vec{y}$, i.e. $f(\vec{x}) \approx y$. This is achieved by \textit{learning} the best parameters $\vec{\theta}$ such that the difference between the model prediction $f(\vec{x}; \vec{\theta})$ and the label $y$ is minimal. Neural nets usually learn using gradient-based algorithms with the help of backpropagation of errors observed at the output layer \cite{rumelhart1985learning}.\\
\indent Following this design, neural networks have achieved state-of-the-art records in several tasks such as image classification \cite{krizhevsky2012imagenet, he2016deep}, image generation \cite{goodfellow2014generative, radford2015unsupervised, zhu2017unpaired}, audio synthesis \cite{oord2016wavenet}, and image captioning \cite{vinyals2015show, xu2015show} among others. However, this dominance was not always the case. Before its resurgence in 2012 by winning the ImageNet Challenge \cite{deng2009imagenet}, training neural networks was notoriously difficult.\\
\indent The difficulty was due to a number of issues, e.g. back in the '80s and '90s, both computing power and training data were not sufficient to harness the full potential of neural networks \cite{goodfellow2016deep}. To a large extent, this was because neural networks were sensitive to initial weights \cite{glorot2010understanding}, and they tended to prematurely stop learning as their gradients decrease to infinitesimally small values due to any or both of the following reasons: (1) their activation functions have small range of values \cite{goodfellow2016deep, maas2013rectifier, nair2010rectified}, and (2) their depth \cite{hochreiter2001gradient}.\\
\indent In this article, we focus on the effect of activation functions to the learning of a neural network. That is, when using an activation function like logistic, the gradients during training become infinitesimally small to the point that a neural network would stop learning. The phenomenon is called the \textit{vanishing gradients} problem \cite{hochreiter2001gradient} -- it occurs when we train neural networks with gradient-based algorithms and backpropagation.\\
\indent A number of solutions have been proposed to avoid the above-mentioned problem, e.g. the use of different activation functions \cite{hahnloser2000digital, maas2013rectifier, nair2010rectified}, and the use of residual connections \cite{he2016deep}. In this article, we focus on the use of rectified linear units (ReLU) and its variants compared to the commonly used activation functions, logistic and tangent.\\
\indent A critical objective of this paper is to formally trace the intellectual lineage of the ReLU activation function, and to correct a prevalent mis-citation of the original version of this paper in contemporary literature. The mathematical foundation of piecewise linear activation functions can be traced back to early biological network modeling by \cite{householder1941theory} and visual feature extraction by \cite{fukushima1969visual}. \cite{hahnloser2000digital} later provided a strong biological motivation for these functions in dynamical networks. However, the seminal integration of ReLU into modern deep learning was established by \cite{nair2010rectified}. Their work formally demonstrated that ReLUs alleviate optimization problems and preserve relative intensity information across multiple layers. \cite{glorot2011deep} subsequently proved that ReLU prevents the vanishing gradient problem in deep feedforward networks.

The original version of this paper has been frequently and erroneously cited as the origin of the ReLU activation function. We must formally clarify that the 2018 study explored an architectural modification using ReLU explicitly at the classification layer in place of the Softmax function. It did not introduce ReLU as a hidden layer activation function. The present study serves to admit this distinction, rectify the literature record, and return attribution to \cite{nair2010rectified} while providing a robust empirical baseline for activation function performance.

\section{Methods}
\subsection{Hardware and Software Setup}
All experiments in this study were conducted in a machine with the following specifications: Intel Core Ultra 7 265 (20) @ 5.300GHz, 48GB DDR5 RAM, and Nvidia GeForce GTX 5060 Ti 16GB GDDR7. Model implementations were written in PyTorch \cite{paszke2019pytorch}, and are available at \href{https://gitlab.com/afagarap/dl-relu.git}{GitLab}. For reproducibility, the seed value used in the experiments was 42.

\subsection{Datasets and Preprocessing}
We evaluated the activation functions across three distinct tasks. For image classification, we utilized the MNIST \cite{lecun1998gradient}, FashionMNIST \cite{xiao2017fashion}, CIFAR10 \cite{krizhevsky2009learning}, and CIFAR100 \cite{krizhevsky2009learning} datasets. Images were resized to $32\times32$ pixels for uniformity of the model architecture. The CIFAR datasets incorporated random crop and horizontal flip augmentations during training. For text classification, we used the IMDB sentiment classification dataset \cite{maas2011learning}. Text data was tokenized using a pre-trained BERT \cite{devlin2019bert} tokenizer with a maximum sequence length of 256 tokens. For image reconstruction, we utilized the MNIST and CIFAR10 datasets resized to $32\times32$ pixels and normalized.

\subsection{Model Architectures}
We engineered generic neural network architectures for each task to isolate the performance of the activation functions. The structural configurations for these models are detailed in Figure \ref{fig:tikz_figure} using block diagrams.

\tikzset{
    layer/.style={rectangle, draw=black, thick, fill=white, text centered, minimum width=3.2cm, minimum height=0.6cm, font=\small\sffamily},
    input/.style={layer, fill=blue!5},
    conv/.style={layer, fill=orange!10},
    pool/.style={layer, fill=green!10},
    fc/.style={layer, fill=red!5},
    embed/.style={layer, fill=yellow!10},
    dim/.style={font=\scriptsize\itshape, text=gray, anchor=west, xshift=0.3cm},
    arrow/.style={-{Stealth[scale=1.2]}, thick},
    title/.style={font=\bfseries\sffamily, yshift=0.5cm}
}

\begin{figure*}[ht]
\centering
\resizebox{\textwidth}{!}{
\begin{tikzpicture}[node distance=0.7cm]

    \node (I1) [input] {Input}; 
    \node [dim] at (I1.east) {$32 \times 32 \times C$};
    \node (T1) [title, above=of I1] {A. Image CNN};
    
    \node (C1) [conv, below=of I1] {Conv $3 \times 3$, 32}; \node [dim] at (C1.east) {$32 \times 32 \times 32$};
    \node (C2) [conv, below=of C1] {Conv $3 \times 3$, 32}; \node [dim] at (C2.east) {$32 \times 32 \times 32$};
    \node (P1) [pool, below=of C2] {Max-Pool $2 \times 2$}; \node [dim] at (P1.east) {$16 \times 16 \times 32$};
    \node (C3) [conv, below=of P1] {Conv $3 \times 3$, 64}; \node [dim] at (C3.east) {$16 \times 16 \times 64$};
    \node (P2) [pool, below=of C3] {Max-Pool $2 \times 2$}; \node [dim] at (P2.east) {$8 \times 8 \times 64$};
    \node (F1) [fc, below=of P2] {Flatten \& Linear}; \node [dim] at (F1.east) {$256$};
    \node (O1) [fc, below=of F1] {Output}; \node [dim] at (O1.east) {$K$};

    \foreach \i/\j in {I1/C1, C1/C2, C2/P1, P1/C3, C3/P2, P2/F1, F1/O1} 
        \draw [arrow] (\i) -- (\j);

    \node (I2) [input, right=5.8cm of I1] {Token Indices}; 
    \node [dim] at (I2.east) {$L \times 1$};
    \node (T2) [title, above=of I2] {B. Text Classifier};

    \node (EM) [embed, below=of I2] {Embedding Layer}; \node [dim] at (EM.east) {$L \times D$};
    \node (TC1) [conv, below=of EM] {Conv1D, 128}; \node [dim] at (TC1.east) {$L \times 128$};
    \node (TP1) [pool, below=of TC1] {Global Max-Pool}; \node [dim] at (TP1.east) {$128$};
    \node (TF1) [fc, below=of TP1] {Dense Layer}; \node [dim] at (TF1.east) {$64$};
    \node (DO) [fc, below=of TF1, fill=gray!10] {Dropout (0.5)}; \node [dim] at (DO.east) {$64$};
    \node (TO) [fc, below=of DO] {Output}; \node [dim] at (TO.east) {$K$};

    \foreach \i/\j in {I2/EM, EM/TC1, TC1/TP1, TP1/TF1, TF1/DO, DO/TO} 
        \draw [arrow] (\i) -- (\j);

    \node (I3) [input, right=5.8cm of I2] {Input Image}; 
    \node [dim] at (I3.east) {$32 \times 32$};
    \node (T3) [title, above=of I3] {C. Autoencoder};

    \node (E1) [conv, below=of I3] {Encoder 1 ($s=2$)}; \node [dim] at (E1.east) {$16 \times 16 \times 16$};
    \node (E2) [conv, below=of E1] {Encoder 2 ($s=2$)}; \node [dim] at (E2.east) {$8 \times 8 \times 32$};
    \node (E3) [conv, below=of E2] {Encoder 3 ($s=2$)}; \node [dim] at (E3.east) {$4 \times 4 \times 64$};
    \node (Z)  [fc, below=of E3, fill=purple!10] {Latent Space}; \node [dim] at (Z.east) {$1024$};
    \node (D1) [conv, below=of Z] {Decoder 1 ($s=2$)}; \node [dim] at (D1.east) {$8 \times 8 \times 32$};
    \node (D2) [conv, below=of D1] {Decoder 2 ($s=2$)}; \node [dim] at (D2.east) {$16 \times 16 \times 16$};
    \node (D3) [conv, below=of D2] {Decoder 3 ($s=2$)}; \node [dim] at (D3.east) {$32 \times 32 \times C$};

    \foreach \i/\j in {I3/E1, E1/E2, E2/E3, E3/Z, Z/D1, D1/D2, D2/D3} 
        \draw [arrow] (\i) -- (\j);

\end{tikzpicture}
}
\caption{Model architectures used for the image classification, text classification, and image reconstruction tasks: (A) Image CNN, (B) Text Classifier, and (C) Autoencoder.}
\label{fig:tikz_figure}
\end{figure*}

The image classification task employed a Convolutional Neural Network (CNN) featuring a VGG style feature extractor. The text classification task utilized a 1D CNN (TextCNN) designed to extract hierarchical linguistic features. The image reconstruction task utilized a convolutional autoencoder with symmetric stride based downsampling and upsampling. Each model was trained independently using ReLU, Hyperbolic Tangent (Tanh), and Logistic (Sigmoid) activation functions in all hidden layers.

\subsection{Experimental Setup and Evaluation Metrics}
To ensure robust statistical significance and computational efficiency, we conducted 10 independent training runs utilizing distinct random seeds for model initialization and data shuffling. All datasets were partitioned using a strict split where 50\% of the total data was allocated for training and 5\% for validation. The remaining 45\% of the training data was discarded to accelerate the experimental pipeline. A completely withheld test set was used exclusively for final evaluation and remained untouched across all configurations.

Training utilized the OneCycle learning rate policy \cite{smith2019super} to achieve super convergence. Image classification models employed Stochastic Gradient Descent (SGD) with a maximum learning rate of $1e-1$. Text classification and image reconstruction models utilized the AdamW optimizer with maximum learning rates of $0.01$ and $0.003$ respectively. Lastly, we  leveraged Automatic Mixed Precision (AMP) using the bfloat16 format for hardware acceleration.

Evaluation metrics for classification tasks included Macro Precision, Macro Recall, and Macro F1 Score in addition to standard accuracy. Reconstruction tasks were evaluated using Mean Squared Error (MSE), Mean Absolute Error (MAE), and Peak Signal to Noise Ratio (PSNR). All final performance metrics are reported as the mean and standard deviation across the 10 independent trials. Statistical significance among the evaluated activation functions was determined using the non parametric Kruskal Wallis H test.

\section{Results}
In the following experiments, we evaluate the empirical performance of the ReLU, Hyperbolic Tangent (Tanh), and Logistic (Sigmoid) activation functions across three distinct tasks.

\subsection{Image Classification}
The image classification models were evaluated on the MNIST, FashionMNIST, CIFAR10, and CIFAR100 datasets. Table \ref{tab:image_classifier_results} summarizes the test loss, accuracy, precision, recall, and F1 Score for each configuration.

\begin{table*}[ht]
\centering
\caption{Mean classification accuracy and standard deviation of the ReLU, Tanh, and Sigmoid activation functions across four image classification datasets over 10 randomized trials. Statistical significance among the activations was determined using the Kruskal Wallis H test ($p < 0.05$).}
\label{tab:image_classifier_results}
\begin{tabular}{|l|l|l|l|l|l|l|l|}
\hline
\rowcolor[HTML]{EFEFEF} 
{\color[HTML]{1F1F1F} \textbf{Dataset}}      & {\color[HTML]{1F1F1F} \textbf{ReLU}} & {\color[HTML]{1F1F1F} \textbf{Tanh}} & {\color[HTML]{1F1F1F} \textbf{Sigmoid}} & {\color[HTML]{1F1F1F} \textbf{H}} & {\color[HTML]{1F1F1F} \textbf{p-value}} & {\color[HTML]{1F1F1F} \textbf{Alpha ($\alpha$)}} & {\color[HTML]{1F1F1F} \textbf{Significant?}} \\ \hline
{\color[HTML]{1F1F1F} \textbf{MNIST}}        & {\color[HTML]{1F1F1F} 99.22 ± 0.08}  & {\color[HTML]{1F1F1F} 99.04 ± 0.06}  & {\color[HTML]{1F1F1F} 11.35 ± 0.00}     & {\color[HTML]{1F1F1F} 25.83}                     & {\color[HTML]{1F1F1F} 2.44e-06}         & {\color[HTML]{1F1F1F} 0.05}               & {\color[HTML]{1F1F1F} Yes}                   \\ \hline
{\color[HTML]{1F1F1F} \textbf{FMNIST}} & {\color[HTML]{1F1F1F} 91.35 ± 0.16}  & {\color[HTML]{1F1F1F} 91.37 ± 0.16}  & {\color[HTML]{1F1F1F} 10.00 ± 0.00}     & {\color[HTML]{1F1F1F} 20.25}                     & {\color[HTML]{1F1F1F} 4.01e-05}         & {\color[HTML]{1F1F1F} 0.05}               & {\color[HTML]{1F1F1F} Yes}                   \\ \hline
{\color[HTML]{1F1F1F} \textbf{CIFAR10}}      & {\color[HTML]{1F1F1F} 79.72 ± 1.34}  & {\color[HTML]{1F1F1F} 79.60 ± 0.15}  & {\color[HTML]{1F1F1F} 10.00 ± 0.00}     & {\color[HTML]{1F1F1F} 21.92}                     & {\color[HTML]{1F1F1F} 1.74e-05}         & {\color[HTML]{1F1F1F} 0.05}               & {\color[HTML]{1F1F1F} Yes}                   \\ \hline
{\color[HTML]{1F1F1F} \textbf{CIFAR100}}     & {\color[HTML]{1F1F1F} 41.91 ± 0.65}  & {\color[HTML]{1F1F1F} 48.53 ± 0.31}  & {\color[HTML]{1F1F1F} 1.00 ± 0.00}      & {\color[HTML]{1F1F1F} 26.79}                     & {\color[HTML]{1F1F1F} 1.52e-06}         & {\color[HTML]{1F1F1F} 0.05}               & {\color[HTML]{1F1F1F} Yes}                   \\ \hline
\end{tabular}
\end{table*}

\begin{figure}
    \centering
    \begin{subfigure}[b]{0.35\textwidth}
        \centering
        \includegraphics[width=\linewidth]{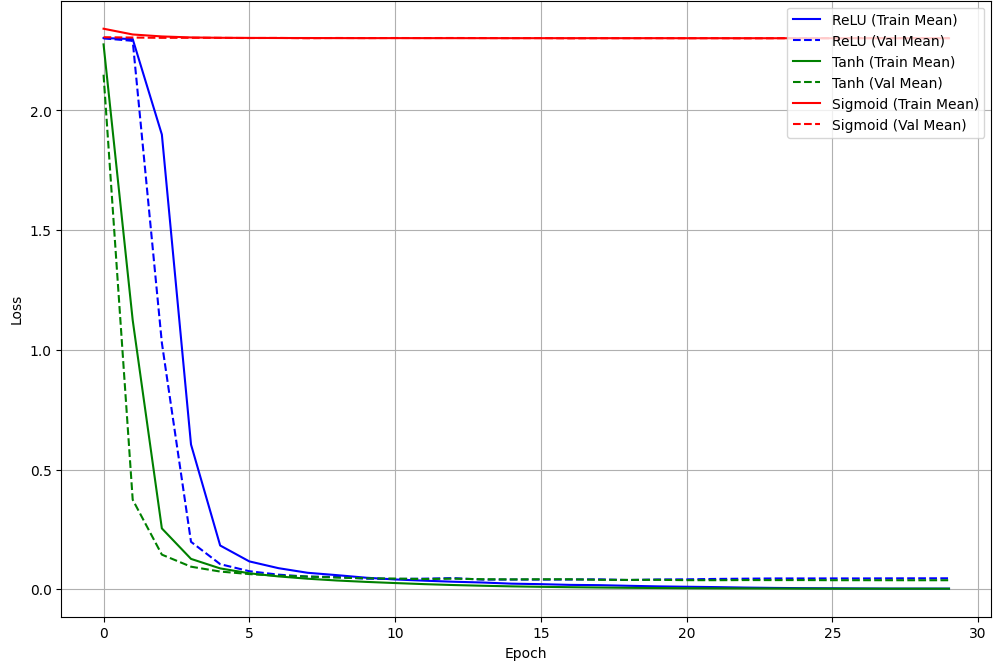}
        \label{fig:mnist_loss}
    \end{subfigure}
    \hspace{5pt} 
    \begin{subfigure}[b]{0.35\textwidth}
        \centering
        \includegraphics[width=\linewidth]{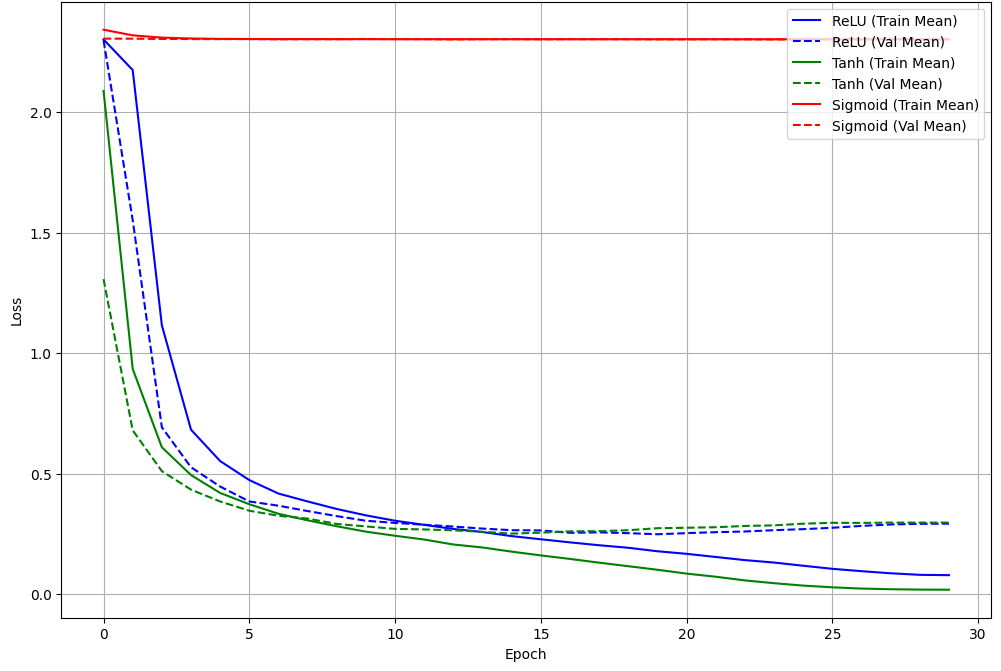}
        \label{fig:fmnist_loss}
    \end{subfigure}

    \vspace{2pt} 
    
    \begin{subfigure}[b]{0.35\textwidth}
        \centering
        \includegraphics[width=\linewidth]{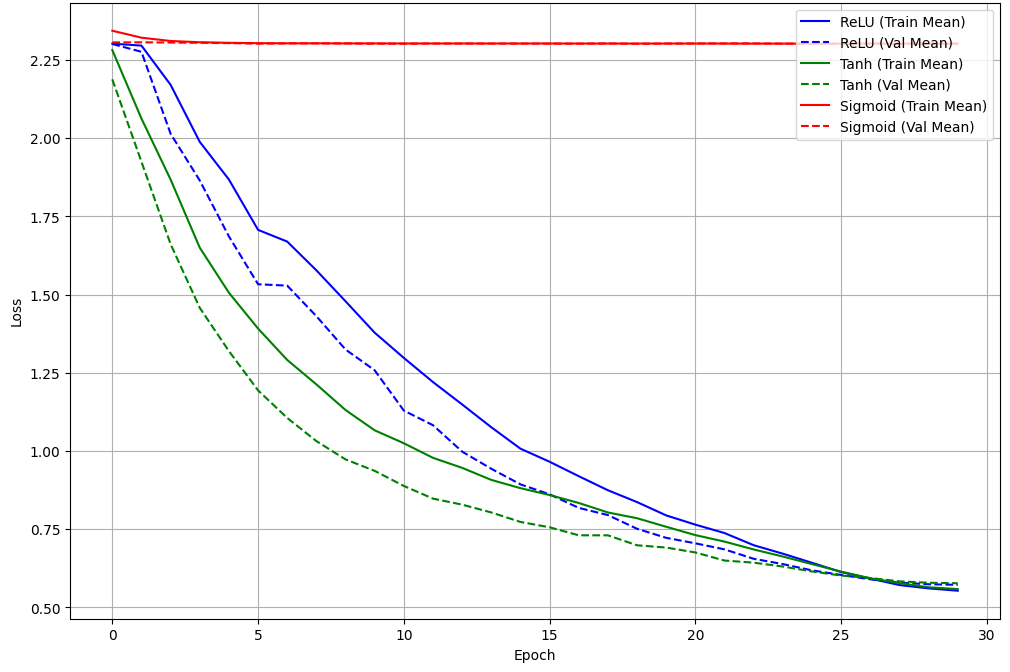}
        \label{fig:cifar10_loss}
    \end{subfigure}
    \hspace{5pt}
    \begin{subfigure}[b]{0.35\textwidth}
        \centering
        \includegraphics[width=\linewidth]{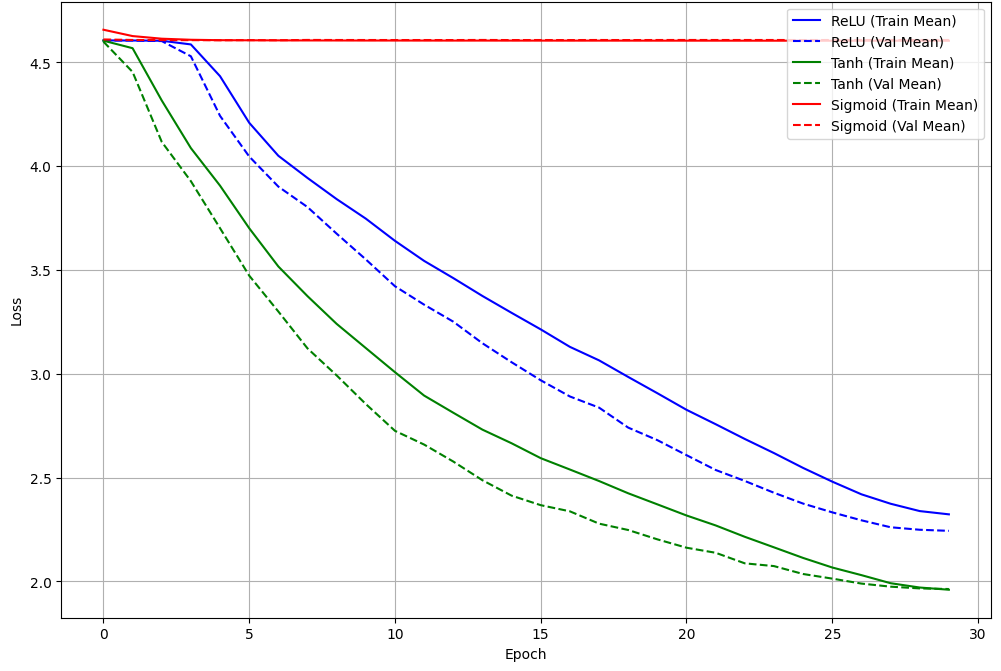}
        \label{fig:cifar100_loss}
    \end{subfigure}
    
    \caption{Training and validation loss curves on MNIST, FashionMNIST, CIFAR10, and CIFAR100 datasets over 10 runs.}
    \label{fig:image_loss_curves}
\end{figure}

The results show models trained with the Logistic (Sigmoid) function were suffering from the vanishing gradient problem \cite{hochreiter2001gradient}. As observed in the loss curves for MNIST, FashionMNIST, CIFAR10, and CIFAR100 in Figure \ref{fig:image_loss_curves}, models utilizing Sigmoid failed to converge entirely. These networks yielded final accuracies equivalent to random probability i.e. roughly 10\% for the 10-class datasets and 1\% for the 100-class dataset. This optimization failure occurs because the derivative of the Sigmoid function reaches a maximum of only 0.25. During backpropagation through multiple convolutional layers, these fractional gradients multiply and exponentially decay to zero. Consequently, the initial layers of the network receive no meaningful error signals and fail to update their weights, preventing the extraction of foundational image features.

\subsection{Text Classification}
\begin{figure}[h]
    \centering
    \includegraphics[width=0.5\linewidth]{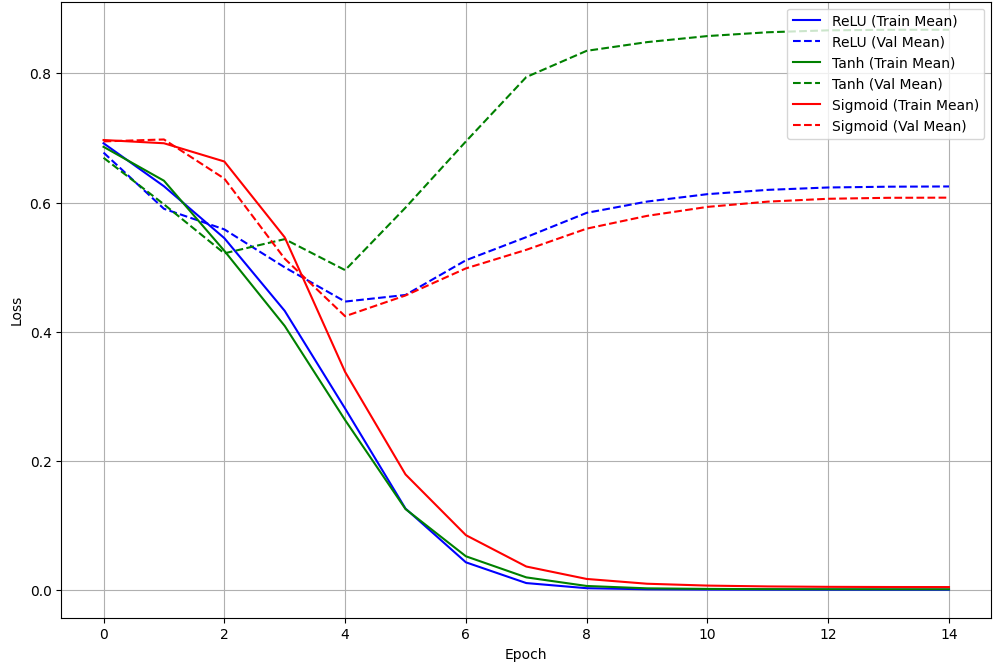}
    \caption{Training and validation loss curves on IMDB dataset.}
    \label{fig:imdb_loss}
\end{figure}

The empirical results from the natural language processing domain present a significantly different convergence behavior compared to the image classification benchmarks. As detailed in Table \ref{tab:text_classifier_results} and visually corroborated by the average loss curves (Figure \ref{fig:imdb_loss}), the Logistic (Sigmoid) and Hyperbolic Tangent (Tanh) functions successfully minimized training loss alongside ReLU. This initial convergence may be attributable to the architectural differences between the models. The 1D Convolutional Neural Network utilized for text classification is substantially shallower than the VGG style feature extractor used for the vision tasks. With fewer hidden layers separating the classification output from the embedding layer, the gradients undergo fewer fractional multiplications during backpropagation. Consequently, the vanishing gradient problem was insufficient to prevent the network from initially learning basic linguistic representations, e.g. sentiment polarity.

\begin{table*}[h]
\centering
\caption{Mean classification accuracy and standard deviation of the ReLU, Tanh, and Sigmoid activation functions on the IMDB sentiment classification dataset over 10 randomized trials. Statistical significance among the activations was determined using the Kruskal Wallis H test (p \textless 0.05).}
\label{tab:my-table}
\begin{tabular}{|l|l|l|l|l|l|l|l|}
\hline
\rowcolor[HTML]{EFEFEF} 
{\color[HTML]{1F1F1F} \textbf{Dataset}} & {\color[HTML]{1F1F1F} \textbf{ReLU}} & {\color[HTML]{1F1F1F} \textbf{Tanh}} & {\color[HTML]{1F1F1F} \textbf{Sigmoid}} & {\color[HTML]{1F1F1F} \textbf{H}} & {\color[HTML]{1F1F1F} \textbf{p-value}} & {\color[HTML]{1F1F1F} \textbf{Alpha ($\alpha$)}} & {\color[HTML]{1F1F1F} \textbf{Significant?}} \\ \hline
{\color[HTML]{1F1F1F} \textbf{IMDB}}    & {\color[HTML]{1F1F1F} 83.33 $\pm$ 0.39}  & {\color[HTML]{1F1F1F} 78.74 $\pm$ 0.36}  & {\color[HTML]{1F1F1F} 79.90 $\pm$ 0.65}     & {\color[HTML]{1F1F1F} 24.58}      & {\color[HTML]{1F1F1F} 4.60e-06}         & {\color[HTML]{1F1F1F} 0.05}               & {\color[HTML]{1F1F1F} Yes}                   \\ \hline
\end{tabular}
\label{tab:text_classifier_results}
\end{table*}

However, the unified average loss trajectories reveal severe overfitting across all three activation functions as training progresses. The validation loss for all configurations reaches its global minimum early in the training cycle (around epoch 3) before diverging significantly, while the training loss continues its asymptotic descent toward zero. Tanh exhibits the most aggressive degradation, with its validation loss climbing steeply after the third epoch. Sigmoid and ReLU also experience validation loss degradation, though their trajectories are smoother than Tanh. This indicates that the 1D CNN rapidly memorizes the training distribution and fails to generalize to unseen data in later epochs, strongly suggesting that the implementation of an early stopping mechanism would yield higher optimal performance for all configurations.

Despite this shared overfitting behavior, ReLU maintained a statistically significant performance advantage at the conclusion of the 15-epoch training cycle, achieving the highest mean accuracy of 83.33\%. The Kruskal Wallis H test confirmed this variance is highly significant ($p < 0.05$). The sustained superiority of ReLU in this domain reinforces its fundamental mathematical advantage. By maintaining a constant gradient of 1.0 in the positive domain, ReLU ensures optimal error signal propagation. Furthermore, the zero bounding of negative inputs naturally induces activation sparsity. This sparsity is highly effective in text classification, as it isolates distinct hierarchical linguistic features and prevents dense, entangled representations that exacerbate overfitting. Ultimately, these results demonstrate that while shallow networks can mathematically tolerate saturating activation functions, ReLU remains the most robust and performant choice across diverse computational domains i.e. computer vision and natural language processing.

\subsection{Image Reconstruction}
A generic convolutional autoencoder was trained to reconstruct inputs from the MNIST and CIFAR10 datasets. Table 3 details the Mean Squared Error (MSE), Mean Absolute Error (MAE), and Peak Signal to Noise Ratio (PSNR).

\begin{table*}[h]
\centering
\caption{Comparative evaluation of activation functions showing the superiority of Tanh over Sigmoid and ReLU on image reconstruction for MNIST and CIFAR10.}
\label{tab:image_reconstruction_results}
\begin{tabular}{|l|l|l|l|l|}
\hline
\rowcolor[HTML]{EFEFEF} 
{\color[HTML]{1F1F1F} \textbf{Dataset}} & {\color[HTML]{1F1F1F} \textbf{Activation}} & {\color[HTML]{1F1F1F} \textbf{MSE}} & {\color[HTML]{1F1F1F} \textbf{MAE}} & {\color[HTML]{1F1F1F} \textbf{PSNR (dB)}} \\ \hline
\multirow{3}{*}{{\color[HTML]{1F1F1F} \textbf{MNIST}}}   & {\color[HTML]{1F1F1F} ReLU}    & {\color[HTML]{1F1F1F} 0.000616}          & {\color[HTML]{1F1F1F} \textbf{0.010673}} & {\color[HTML]{1F1F1F} 32.11}          \\ \cline{2-5} 
                                                         & {\color[HTML]{1F1F1F} Tanh}    & {\color[HTML]{1F1F1F} \textbf{0.000479}} & {\color[HTML]{1F1F1F} 0.011207}          & {\color[HTML]{1F1F1F} \textbf{33.19}} \\ \cline{2-5} 
                                                         & {\color[HTML]{1F1F1F} Sigmoid} & {\color[HTML]{1F1F1F} 0.061017}          & {\color[HTML]{1F1F1F} 0.152161}          & {\color[HTML]{1F1F1F} 12.15}          \\ \hline
\multirow{3}{*}{{\color[HTML]{1F1F1F} \textbf{CIFAR10}}} & {\color[HTML]{1F1F1F} ReLU}    & {\color[HTML]{1F1F1F} 0.002873}          & {\color[HTML]{1F1F1F} 0.038310}          & {\color[HTML]{1F1F1F} 25.42}          \\ \cline{2-5} 
                                                         & {\color[HTML]{1F1F1F} Tanh}    & {\color[HTML]{1F1F1F} \textbf{0.001085}} & {\color[HTML]{1F1F1F} \textbf{0.023783}} & {\color[HTML]{1F1F1F} \textbf{29.64}} \\ \cline{2-5} 
                                                         & {\color[HTML]{1F1F1F} Sigmoid} & {\color[HTML]{1F1F1F} 0.005875}          & {\color[HTML]{1F1F1F} 0.055561}          & {\color[HTML]{1F1F1F} 22.31}          \\ \hline
\end{tabular}
\end{table*}

Higher PSNR values indicate superior image fidelity. Tanh provided the most accurate reconstructions for both datasets, achieving 33.19 dB on MNIST and 29.64 dB on CIFAR10. ReLU followed closely. The Sigmoid function yielded the lowest reconstruction quality, matching its poor performance in the image classification benchmarks.

\begin{figure}
    \centering
    \begin{subfigure}[b]{0.45\linewidth}
        \centering
        \includegraphics[width=\linewidth]{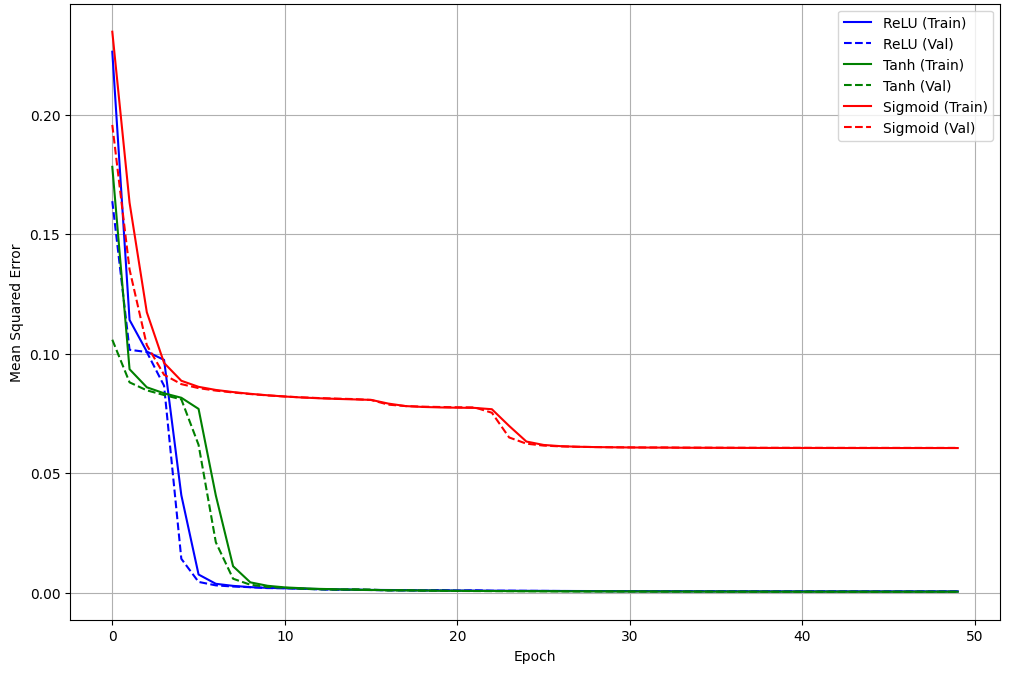}
        \label{fig:mnist_mse}    
    \end{subfigure}
    \begin{subfigure}[b]{0.45\linewidth}
        \centering
        \includegraphics[width=\linewidth]{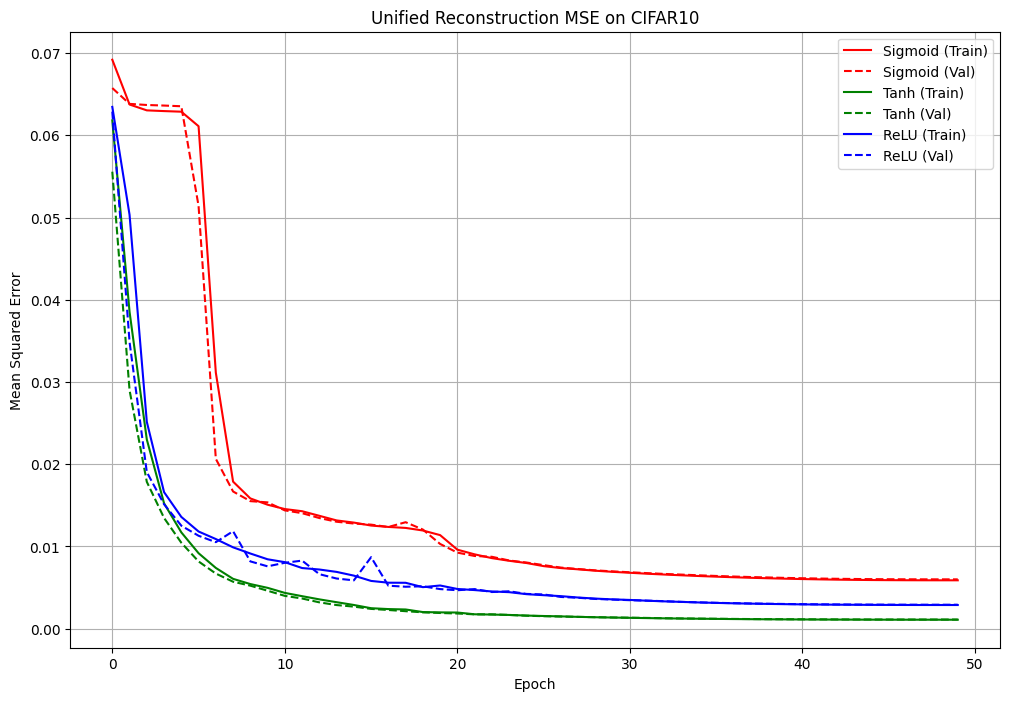}
        \label{fig:cifar10_mse}    
    \end{subfigure}
\end{figure}

\begin{figure}
    \centering
    \begin{subfigure}[b]{0.3\linewidth}
        \centering
        \includegraphics[width=\linewidth]{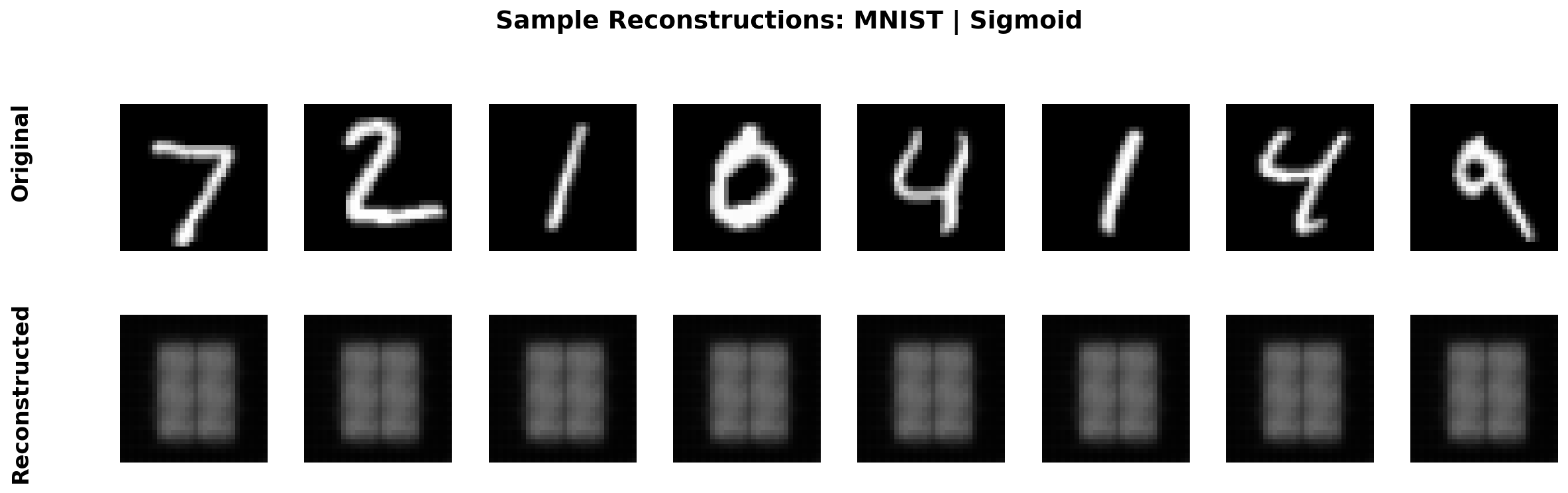}
        \label{fig:sigmoid_mnist_reconstructions}
    \end{subfigure}
    \begin{subfigure}[b]{0.3\linewidth}
        \centering
        \includegraphics[width=\linewidth]{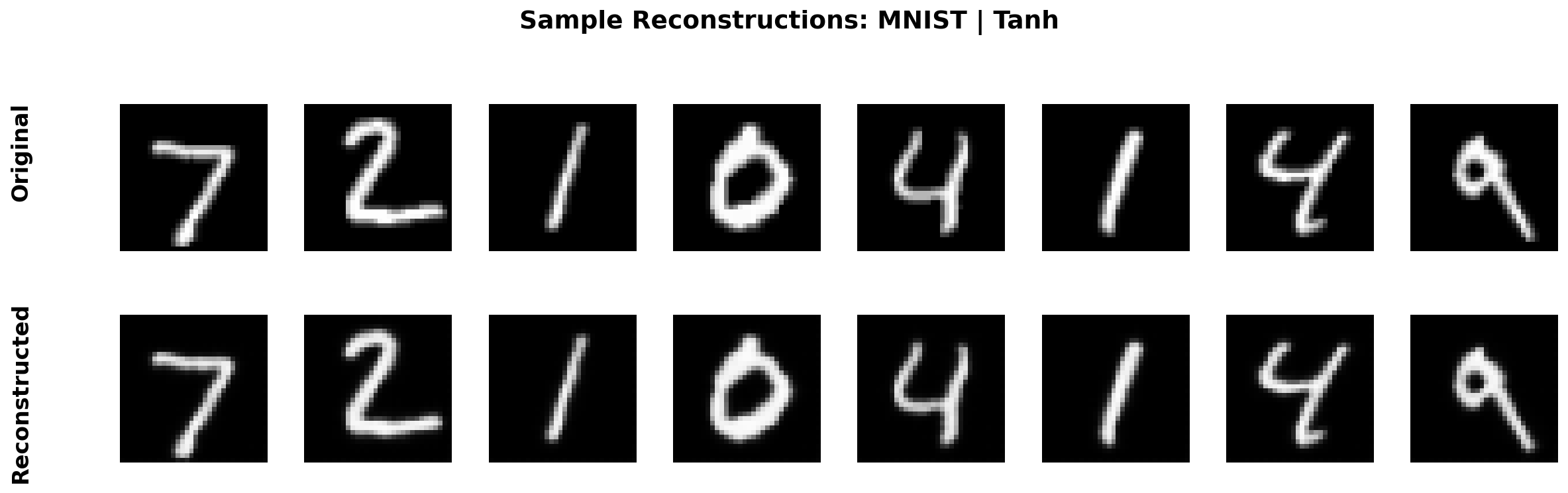}
        \label{fig:tanh_mnist_reconstructions}
    \end{subfigure}
    \begin{subfigure}[b]{0.3\linewidth}
        \centering
        \includegraphics[width=\linewidth]{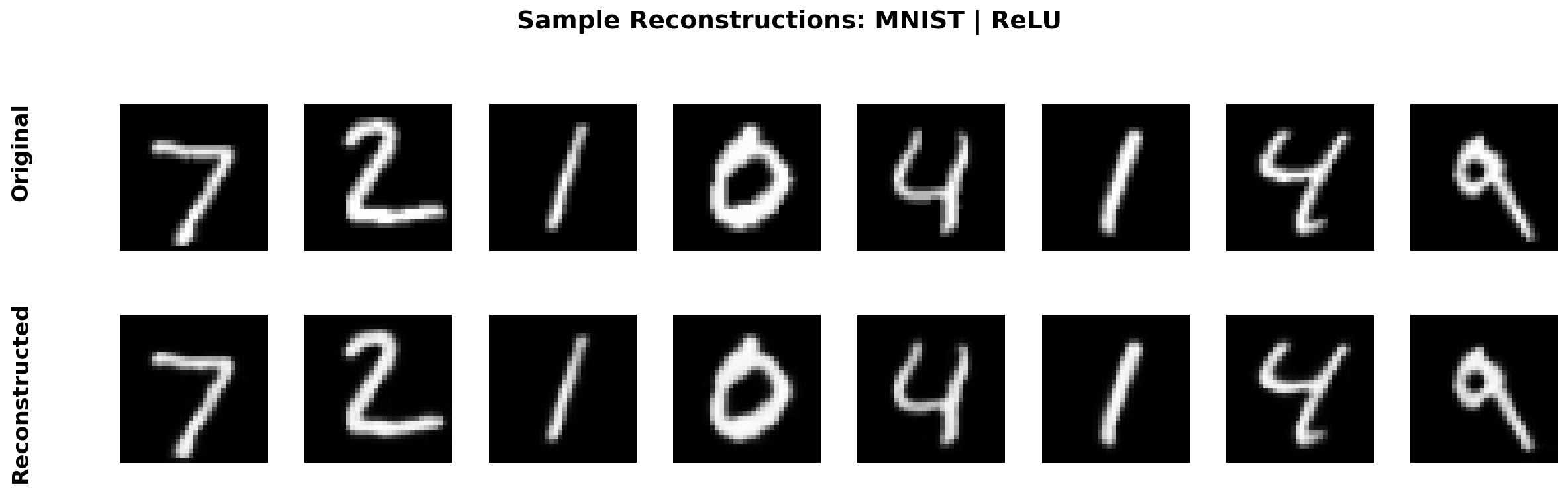}
        \label{fig:relu_mnist_reconstructions}
    \end{subfigure}
    \vspace{5pt}
    \begin{subfigure}[b]{0.3\linewidth}
        \centering
        \includegraphics[width=\linewidth]{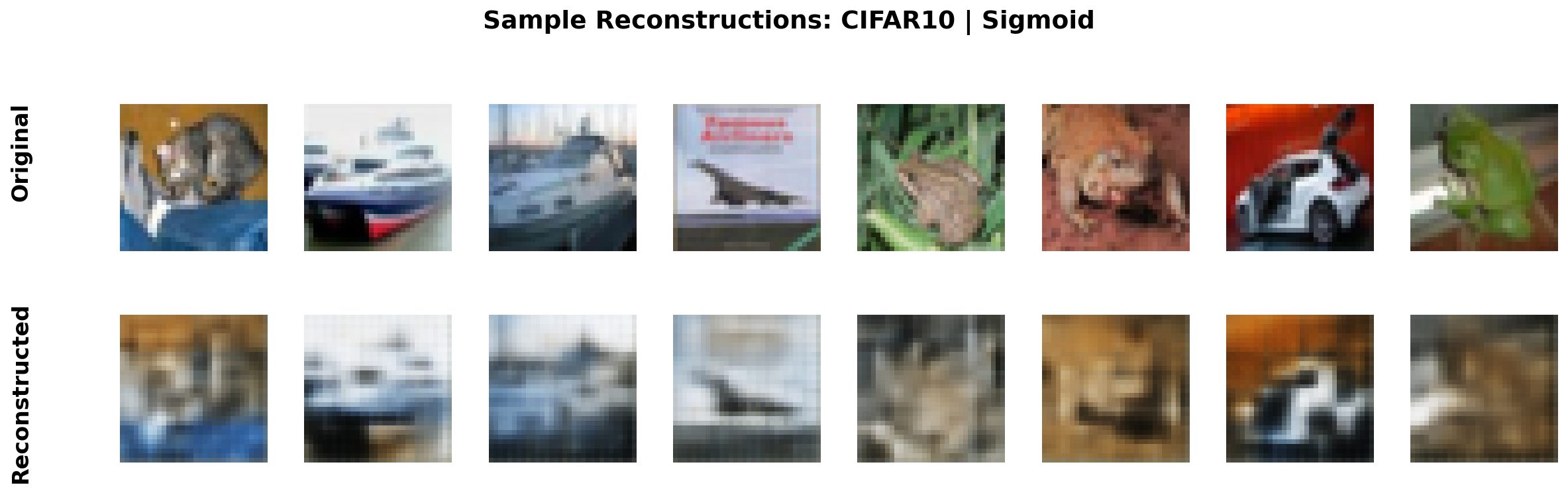}
        \label{fig:sigmoid_cifar10_reconstructions}
    \end{subfigure}
    \begin{subfigure}[b]{0.3\linewidth}
        \centering
        \includegraphics[width=\linewidth]{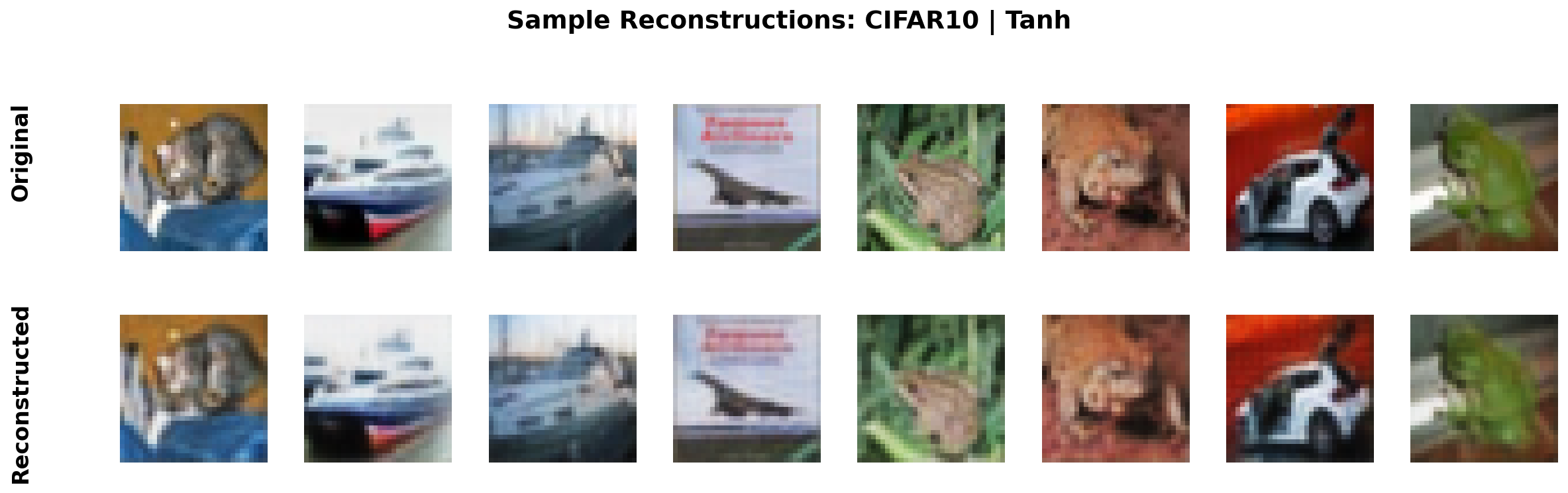}
        \label{fig:tanh_cifar10_reconstructions}
    \end{subfigure}
    \begin{subfigure}[b]{0.3\linewidth}
        \centering
        \includegraphics[width=\linewidth]{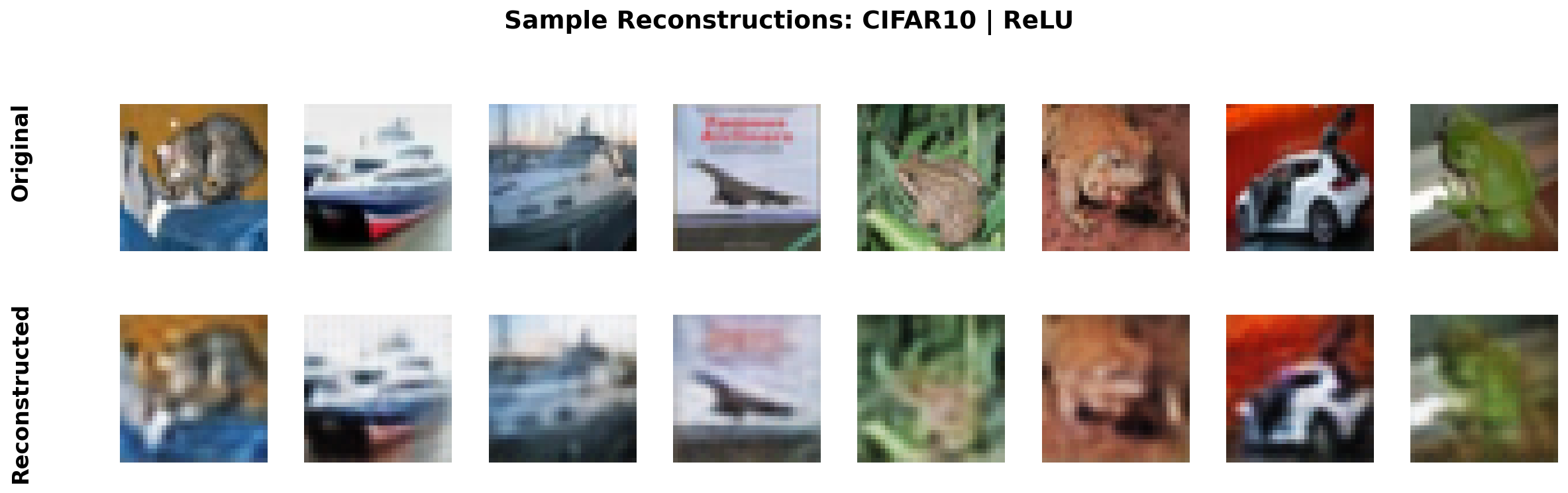}
        \label{fig:relu_cifar10_reconstructions}
    \end{subfigure}
    \caption{Sample image reconstructions for the MNIST dataset (top) and CIFAR10 dataset (bottom) using Sigmoid, ReLU, and Tanh activation functions.}
\end{figure}

\subsection{Discussion}
The empirical results confirm the theoretical limitations of the Logistic function in deep architectures. Sigmoid saturates at its extremes and causes gradients to vanish rapidly during backpropagation. This prevents the initial layers of deep convolutional networks from learning meaningful feature representations. Tanh also suffers from saturation but mitigates the issue slightly by being zero centered. ReLU avoids positive saturation entirely. This non saturating property allows gradients to flow through deep architectures efficiently.

\section{Conclusion}
We conducted an empirical comparison of the Rectified Linear Unit against the Logistic and Hyperbolic Tangent activation functions across image classification, text classification, and image reconstruction tasks. The results validate that ReLU and Tanh consistently outperform Sigmoid in deep neural networks. Sigmoid suffers from severe gradient vanishing in convolutional architectures.

Furthermore, this paper formally clarifies the intellectual lineage of ReLU. The function was integrated into deep learning by Nair and Hinton (2010). Recent literature misattributing the activation function to Agarap (2018) is incorrect, as that specific study only investigated ReLU at the classification layer. This document serves to correct the citation record and reaffirm the foundational contributions of early biological and computational researchers to modern deep learning.

\newpage
\begin{appendices}
    \section{Supplemental Results}
        The extended classification metrics detailed in the preceding tables corroborate the primary accuracy findings while highlighting the stability of the models across the 10 randomized trials. For the image classification tasks, the precision, recall, and F1 scores exhibit remarkably low standard deviations for both ReLU and Tanh (frequently below 0.01). This indicates highly stable convergence irrespective of the initial weight distributions. Conversely, Sigmoid consistently yields F1 scores near zero with zero variance. This mathematically reflects a complete optimization failure, where the model outputs predictions equivalent to random guessing i.e. 10\% recall for 10-class datasets and 1\% recall for the 100-class dataset. Notably, Tanh marginally outperforms ReLU across all three extended metrics on the CIFAR100 dataset, suggesting its zero-centered nature may offer slight representational advantages in highly multiplexed visual feature spaces.
        
        \begin{table*}[h]
\centering
\caption{Mean precision, recall, and F1 score with standard deviation for the ReLU, Tanh, and Sigmoid activation functions across four image classification datasets over 10 randomized trials.}
\label{tab:image_classification_metrics}
\begin{tabular}{|l|l|l|l|l|}
\hline
\rowcolor[HTML]{EFEFEF} 
{\color[HTML]{1F1F1F} \textbf{Dataset}} & {\color[HTML]{1F1F1F} \textbf{Activation}} & {\color[HTML]{1F1F1F} \textbf{Precision}} & {\color[HTML]{1F1F1F} \textbf{Recall}} & {\color[HTML]{1F1F1F} \textbf{F1-Score}} \\ \hline
\multirow{3}{*}{{\color[HTML]{1F1F1F} \textbf{MNIST}}}        & {\color[HTML]{1F1F1F} ReLU}    & {\color[HTML]{1F1F1F} 0.9922 $\pm$ 0.0008} & {\color[HTML]{1F1F1F} 0.9922 $\pm$ 0.0008} & {\color[HTML]{1F1F1F} 0.9922 $\pm$ 0.0008} \\ \cline{2-5} 
                                                              & {\color[HTML]{1F1F1F} Tanh}    & {\color[HTML]{1F1F1F} 0.9903 $\pm$ 0.0006} & {\color[HTML]{1F1F1F} 0.9903 $\pm$ 0.0006} & {\color[HTML]{1F1F1F} 0.9903 $\pm$ 0.0006} \\ \cline{2-5} 
                                                              & {\color[HTML]{1F1F1F} Sigmoid} & {\color[HTML]{1F1F1F} 0.0114 $\pm$ 0.0000} & {\color[HTML]{1F1F1F} 0.1000 $\pm$ 0.0000} & {\color[HTML]{1F1F1F} 0.0204 $\pm$ 0.0000} \\ \hline
\multirow{3}{*}{{\color[HTML]{1F1F1F} \textbf{FashionMNIST}}} & {\color[HTML]{1F1F1F} ReLU}    & {\color[HTML]{1F1F1F} 0.9134 $\pm$ 0.0017} & {\color[HTML]{1F1F1F} 0.9135 $\pm$ 0.0016} & {\color[HTML]{1F1F1F} 0.9134 $\pm$ 0.0017} \\ \cline{2-5} 
                                                              & {\color[HTML]{1F1F1F} Tanh}    & {\color[HTML]{1F1F1F} 0.9134 $\pm$ 0.0015} & {\color[HTML]{1F1F1F} 0.9137 $\pm$ 0.0016} & {\color[HTML]{1F1F1F} 0.9134 $\pm$ 0.0015} \\ \cline{2-5} 
                                                              & {\color[HTML]{1F1F1F} Sigmoid} & {\color[HTML]{1F1F1F} 0.0100 $\pm$ 0.0000} & {\color[HTML]{1F1F1F} 0.1000 $\pm$ 0.0000} & {\color[HTML]{1F1F1F} 0.0182 $\pm$ 0.0000} \\ \hline
\multirow{3}{*}{{\color[HTML]{1F1F1F} \textbf{CIFAR10}}}      & {\color[HTML]{1F1F1F} ReLU}    & {\color[HTML]{1F1F1F} 0.7969 $\pm$ 0.0135} & {\color[HTML]{1F1F1F} 0.7972 $\pm$ 0.0134} & {\color[HTML]{1F1F1F} 0.7968 $\pm$ 0.0136} \\ \cline{2-5} 
                                                              & {\color[HTML]{1F1F1F} Tanh}    & {\color[HTML]{1F1F1F} 0.7961 $\pm$ 0.0013} & {\color[HTML]{1F1F1F} 0.7960 $\pm$ 0.0015} & {\color[HTML]{1F1F1F} 0.7958 $\pm$ 0.0013} \\ \cline{2-5} 
                                                              & {\color[HTML]{1F1F1F} Sigmoid} & {\color[HTML]{1F1F1F} 0.0100 $\pm$ 0.0000} & {\color[HTML]{1F1F1F} 0.1000 $\pm$ 0.0000} & {\color[HTML]{1F1F1F} 0.0182 $\pm$ 0.0000} \\ \hline
\multirow{3}{*}{{\color[HTML]{1F1F1F} \textbf{CIFAR100}}}     & {\color[HTML]{1F1F1F} ReLU}    & {\color[HTML]{1F1F1F} 0.4124 $\pm$ 0.0068} & {\color[HTML]{1F1F1F} 0.4191 $\pm$ 0.0065} & {\color[HTML]{1F1F1F} 0.4101 $\pm$ 0.0065} \\ \cline{2-5} 
                                                              & {\color[HTML]{1F1F1F} Tanh}    & {\color[HTML]{1F1F1F} 0.4802 $\pm$ 0.0027} & {\color[HTML]{1F1F1F} 0.4853 $\pm$ 0.0031} & {\color[HTML]{1F1F1F} 0.4785 $\pm$ 0.0034} \\ \cline{2-5} 
                                                              & {\color[HTML]{1F1F1F} Sigmoid} & {\color[HTML]{1F1F1F} 0.0001 $\pm$ 0.0000} & {\color[HTML]{1F1F1F} 0.0100 $\pm$ 0.0000} & {\color[HTML]{1F1F1F} 0.0002 $\pm$ 0.0000} \\ \hline
\end{tabular}
\end{table*}
        
        In the text classification domain, the tight standard deviations (e.g. $\pm 0.0039$ for ReLU) confirm that the experimental protocol yielded statistically robust outcomes. Consistent with the accuracy results, ReLU achieves the highest F1 score (0.8333). Interestingly, Sigmoid outperforms Tanh in precision, recall, and F1 score within this specific natural language context. However, across both deep vision architectures and shallower text classifiers, the comprehensive metric suite reaffirms that ReLU provides the most optimal, stable, and computationally efficient feature representation.
        
        \begin{table*}[h]
\centering
\caption{Mean precision, recall, and F1 score with standard deviation for the ReLU, Tanh, and Sigmoid activation functions on the IMDB sentiment classification dataset over 10 randomized trials.}
\label{tab:text_classification_metrics}
\begin{tabular}{|l|l|l|l|l|}
\hline
\rowcolor[HTML]{EFEFEF} 
{\color[HTML]{1F1F1F} \textbf{Dataset}} & {\color[HTML]{1F1F1F} \textbf{Activation}} & {\color[HTML]{1F1F1F} \textbf{Precision}} & {\color[HTML]{1F1F1F} \textbf{Recall}} & {\color[HTML]{1F1F1F} \textbf{F1-Score}} \\ \hline
\multirow{3}{*}{{\color[HTML]{1F1F1F} \textbf{IMDB}}} & {\color[HTML]{1F1F1F} ReLU}    & {\color[HTML]{1F1F1F} 0.8333 $\pm$ 0.0039} & {\color[HTML]{1F1F1F} 0.8333 $\pm$ 0.0039} & {\color[HTML]{1F1F1F} 0.8333 $\pm$ 0.0039} \\ \cline{2-5} 
                                                      & {\color[HTML]{1F1F1F} Tanh}    & {\color[HTML]{1F1F1F} 0.7875 $\pm$ 0.0037} & {\color[HTML]{1F1F1F} 0.7874 $\pm$ 0.0036} & {\color[HTML]{1F1F1F} 0.7874 $\pm$ 0.0036} \\ \cline{2-5} 
                                                      & {\color[HTML]{1F1F1F} Sigmoid} & {\color[HTML]{1F1F1F} 0.7992 $\pm$ 0.0066} & {\color[HTML]{1F1F1F} 0.7990 $\pm$ 0.0065} & {\color[HTML]{1F1F1F} 0.7990 $\pm$ 0.0065} \\ \hline
\end{tabular}
\end{table*}
\end{appendices}


\begin{thebibliography}{99}
\bibitem{deng2009imagenet} Deng, Jia, et al. ``Imagenet: A large-scale hierarchical image database.'' 2009 IEEE conference on computer vision and pattern recognition. Ieee, 2009.
\bibitem{devlin2019bert} Devlin, Jacob, et al. ``Bert: Pre-training of deep bidirectional transformers for language understanding.'' Proceedings of the 2019 conference of the North American chapter of the association for computational linguistics: human language technologies, volume 1 (long and short papers). 2019.
\bibitem{fukushima1969visual} Fukushima, Kunihiko. ``Visual Feature Extraction by a Multilayered Network of Analog Threshold Elements.'' IEEE Transactions on Systems Science and Cybernetics, vol. 5, no. 4, 1969, pp. 322-333.
\bibitem{glorot2010understanding} Glorot, Xavier, and Yoshua Bengio. ``Understanding the difficulty of training deep feedforward neural networks.'' Proceedings of the thirteenth international conference on artificial intelligence and statistics. 2010.
\bibitem{glorot2011deep} Glorot, Xavier, Antoine Bordes, and Yoshua Bengio. ``Deep Sparse Rectifier Neural Networks.'' Proceedings of the Fourteenth International Conference on Artificial Intelligence and Statistics, 2011, pp. 315-323.
\bibitem{goodfellow2014generative} Goodfellow, Ian, et al. ``Generative adversarial nets.'' Advances in neural information processing systems. 2014.
\bibitem{goodfellow2016deep} Goodfellow, Ian, Yoshua Bengio, and Aaron Courville. Deep learning. MIT press, 2016.
\bibitem{hahnloser2000digital} Hahnloser, Richard HR, et al. ``Digital selection and analogue amplification coexist in a cortex-inspired silicon circuit.'' Nature 405.6789 (2000): 947.
\bibitem{he2016deep} He, Kaiming, et al. ``Deep residual learning for image recognition.'' Proceedings of the IEEE conference on computer vision and pattern recognition. 2016.
\bibitem{hochreiter2001gradient} Hochreiter, Sepp, et al. ``Gradient flow in recurrent nets: the difficulty of learning long-term dependencies.'' (2001).
\bibitem{householder1941theory} Householder, Alston S. ``A Theory of Steady-State Activity in Nerve-Net.'' The Bulletin of Mathematical Biophysics, vol. 3, no. 2, 1941, pp. 63-69.
\bibitem{krizhevsky2009learning} Krizhevsky, Alex, and Geoffrey Hinton. ``Learning multiple layers of features from tiny images.'' (2009): 7.
\bibitem{krizhevsky2012imagenet} Krizhevsky, Alex, Ilya Sutskever, and Geoffrey E. Hinton. ``Imagenet classification with deep convolutional neural networks.'' Advances in neural information processing systems. 2012.
\bibitem{lecun1998gradient} LeCun, Yann, et al. ``Gradient-based learning applied to document recognition.'' Proceedings of the IEEE 86.11 (1998): 2278-2324.
\bibitem{maas2011learning} Maas, Andrew L., et al. ``Learning word vectors for sentiment analysis.'' Proceedings of the 49th annual meeting of the association for computational linguistics: Human language technologies-volume 1. Association for Computational Linguistics, 2011.
\bibitem{maas2013rectifier} Maas, Andrew L., Awni Y. Hannun, and Andrew Y. Ng. ``Rectifier nonlinearities improve neural network acoustic models.'' Proc. icml. Vol. 30. No. 1. 2013.
\bibitem{nair2010rectified} Nair, Vinod, and Geoffrey E. Hinton. ``Rectified linear units improve restricted boltzmann machines.'' Proceedings of the 27th international conference on machine learning (ICML-10). 2010.
\bibitem{oord2016wavenet} Oord, Aaron van den, et al. ``Wavenet: A generative model for raw audio.'' arXiv preprint arXiv:1609.03499 (2016).
\bibitem{paszke2019pytorch} Paszke, Adam, et al. ``Pytorch: An imperative style, high-performance deep learning library.'' Advances in neural information processing systems 32 (2019).
\bibitem{radford2015unsupervised} Radford, Alec, Luke Metz, and Soumith Chintala. ``Unsupervised representation learning with deep convolutional generative adversarial networks.'' arXiv preprint arXiv:1511.06434 (2015).
\bibitem{rumelhart1985learning} Rumelhart, David E., Geoffrey E. Hinton, and Ronald J. Williams. Learning internal representations by error propagation. No. ICS-8506. California Univ San Diego La Jolla Inst for Cognitive Science, 1985.
\bibitem{smith2019super} Smith, Leslie N., and Nicholay Topin. ``Super-convergence: Very fast training of neural networks using large learning rates.'' Artificial intelligence and machine learning for multi-domain operations applications. Vol. 11006. SPIE, 2019.
\bibitem{vinyals2015show} Vinyals, Oriol, et al. ``Show and tell: A neural image caption generator.'' Proceedings of the IEEE conference on computer vision and pattern recognition. 2015.
\bibitem{xiao2017fashion} Xiao, Han, Kashif Rasul, and Roland Vollgraf. ``Fashion-mnist: a novel image dataset for benchmarking machine learning algorithms.'' arXiv preprint arXiv:1708.07747 (2017).
\bibitem{xu2015show} Xu, Kelvin, et al. ``Show, attend and tell: Neural image caption generation with visual attention.'' International conference on machine learning. 2015.
\bibitem{zhu2017unpaired} Zhu, Jun-Yan, et al. ``Unpaired image-to-image translation using cycle-consistent adversarial networks.'' Proceedings of the IEEE international conference on computer vision. 2017.

\end{thebibliography}
\end{document}